\documentclass[lettersize,journal]{IEEEtran}
\usepackage{amsmath,amsfonts}
\usepackage{algorithmic}
\usepackage{array}
\usepackage[caption=false,font=normalsize,labelfont=sf,textfont=sf]{subfig}
\usepackage{textcomp}
\usepackage{stfloats}
\usepackage{url}
\usepackage{verbatim}
\usepackage{graphicx}
\usepackage{tikz}
\usepackage{cite}
\usepackage{amssymb}
\usepackage{booktabs}
\usepackage{xcolor} 
\usepackage{enumitem} 
\usepackage[most]{tcolorbox}
\usepackage{tcolorbox}
\usepackage{wasysym} 
\usepackage{xspace}
\usepackage{caption}
\usepackage{makecell}
\usepackage{multirow}
\usepackage{booktabs}
\usepackage{diagbox}
\usepackage{colortbl}
\usepackage{float}
\usepackage{booktabs}
\usepackage{listings}
\usepackage{fancyvrb}
\usepackage{subcaption}
\usepackage[ruled,vlined]{algorithm2e}
\usepackage{svg}
\definecolor{mygreen}{RGB}{220,255,220}
\definecolor{myred}{RGB}{255,220,220}
\definecolor{mycyan}{RGB}{220,255,255}
\definecolor{highlight}{RGB}{200,230,255}
\definecolor{worst}{RGB}{255,200,200}

\newcommand{\toolname}{NeuroDeX\xspace}
\newtcolorbox{RQBox}{
    colback=gray!10,     
    colframe=black,    
    arc=5pt,             
    boxrule=0.8pt,       
    left=6pt, right=6pt, 
    top=6pt, bottom=6pt, 
    boxsep=0pt,          
    before upper={\parindent15pt}, 
    fontupper=\normalsize,    
}

\def\BibTeX{{\rm B\kern-.05em{\sc i\kern-.025em b}\kern-.08em
    T\kern-.1667em\lower.7ex\hbox{E}\kern-.125emX}}
\usepackage{balance}
\begin{document}
\title{NeuroDeX: Unlocking Diverse Support in Decompiling Deep Neural Network Executables}
\author{Yilin Li\textsuperscript{1,2}, Guozhu Meng\textsuperscript{1,2,*}, Mingyang Sun\textsuperscript{1,2}, Yanzhong Wang\textsuperscript{1,2}, Kun Sun\textsuperscript{1,2}, Hailong Chang\textsuperscript{1,2}, Yuekang Li\textsuperscript{3}\\
\textsuperscript{1}Institute of Information Engineering, CAS, China \\
\textsuperscript{2}School of Cybersecurity, UCAS, China \\
\textsuperscript{3}University of New South Wales, Australia\\

$ \{ $liyilin2023, mengguozhu, sunmingyang, wangyanzhong, sunkun2023, changhailong$ \}$@iie.ac.cn \\
yuekang.li@unsw.edu.au
\thanks{*Corresponding author.}
}

\maketitle

\begin{abstract}
On-device deep learning models have extensive real-world demands. Deep learning compilers efficiently compile models into executables for deployment on edge devices, but these executables may face the threat of reverse engineering. Previous studies have attempted to decompile DNN executables, but they face challenges in handling compilation optimizations and analyzing quantized compiled models.

In this paper, we present NeuroDeX to unlock diverse support in decompiling DNN executables.
NeuroDeX leverages the semantic understanding capabilities of LLMs along with dynamic analysis to accurately and efficiently perform operator type recognition, operator attribute recovery and model reconstruction. NeuroDeX can recover DNN executables into high-level models towards compilation optimizations, different architectures and quantized compiled models. We conduct experiments on 96 DNN executables across 12 common DNN models. Extensive experimental results demonstrate that NeuroDeX can decompile non-quantized executables into nearly identical high-level models. NeuroDeX can recover functionally  similar high-level models for quantized executables, achieving an average top-1 accuracy of 72\%. NeuroDeX offers a more comprehensive and effective solution compared to previous DNN executables decompilers.
\end{abstract}


\section{Introduction}
In recent years, deep learning (DL) has rapidly advanced in the real world. Deploying deep neural networks (DNNs) on edge devices can meet the real-time requirements of edge computing, enhance privacy protection and enable offline inference capabilities, making DNNs widely applicable in real-world scenarios. 
DL compilers, such as TVM~\cite{chen2018tvm} and GLOW~\cite{rotem2018glow}, can compile high-level DNN models into executables for inference on edge devices. DNNs are composed of different neural network operators (e.g., \textit{Conv}, \textit{Relu}), and DL compilers compiles these operators into operator functions in executables. DL compilers optimize models during compilation to improve inference efficiency and reduce deployment environment dependencies, which provides a good solution for deploying  models on edge devices~\cite{wu2022usage,liu2023deploying,li2020deep}. 

In DNN executables, the operators and weights are compiled into incomprehensible machine code, thereby reducing the risk of model stealing attacks compared to white-box deployment. However, DNN executables may still pose security risks due to decompilation. DNN executables can potentially be decompiled to original high-level models. This undermines the intellectual property of the model owners, especially for models trained on private data. Based on the recovered high-level models, attackers can perform white-box adversarial attacks and backdoor attacks, threatening the secure use of DNN executables. 

Previous works~\cite{zhang2023libsteal,wu2022dnd,liu2023decompiling,shi2024research,wuneuroscope} have attempted to decompile DNN binaries, but the existing methods have certain limitations. They struggle to simultaneously address compilation optimizations, support different architectures, and accommodate quantized compiled model, all of which are essential for meeting the demands of modern DL applications. Previous decompilers also face issues such as insufficient accuracy and reliance on prior knowledge when recognizing operator types, and they also often come with significant analysis overhead.

To address these limitations, we propose \toolname{} to unlock diverse support in decompiling DNN executables.
\toolname{} begins by collecting operator function information including disassembled and decompiled code from DNN executables. Next, \toolname{} utilizes the semantic understanding abilities of LLMs to design an accurate and scalable method for operator type recognition. Subsequently, \toolname{} utilizes dynamic analysis and LLM-based code understanding to finish operator attribute recovery. Finally, \toolname{} reconstructs the model's computational graph and weights through dynamic analysis. 
\toolname{} features an accurate operator type recognition and operator attribute recovery mechanism that does not rely on prior knowledge such as compiler versions or training data. \toolname{} can accurately recover fused operators and its core components do not depend on resource-intensive analysis techniques like symbolic execution, allowing for rapid and efficient analysis. Furthermore, \toolname{} is extendable to different architectures, different DL compilers, and quantized models.

\toolname{} is evaluated on 88 non-quantized DNN executables and \toolname{} can accurately recover them into nearly identical high-level models. \toolname{} adapts for the different compiler versions, accommodates a wider range of models, and supports different architectures. The operator type recognition accuracy for all TVM executables and GLOW executables reaches 99.22\% and 97.62\% respectively. The operator attribute recovery accuracy is nearly 100\%. \toolname{} incorporates robust error fix strategies, and \textbf{all} the recovered model's inference accuracy reaches 100\% after the errors are fixed. Additionally, we evaluate \toolname{} on 8 quantized compiled DNN executables, the results indicate that \toolname{} can successfully recover functionally similar high-level models. For model inference, the average top\_1 accuracy is 72\%, and the average top\_5 accuracy is 86\%.

Our contributions are summarized as follows:
\begin{itemize}
    \item We propose \toolname{} to provide diverse support in decompiling DNN executables. It can decompile DNN executables into high-level models towards different DL compilers, different architectures and quantized compiled models.
    \item We design a mechanism that leverages the semantic understanding capabilities of LLMs along with dynamic analysis to accurately and efficiently perform operator type recognition and operator attribute recovery, overcoming the limitations of previous works.
    \item We conduct experiments on 96 DNN executables across 12 common DNN models. Extensive experimental results demonstrate that \toolname{} can successfully decompile various types of DNN executables, providing more comprehensive support than previous decompilers. 
\end{itemize}

\section{Foundation and Problem Statement}
\subsection{Deep Learning Compiler}
DNNs are computational models consisting of multiple layers. These layers are constructed from different types of neural network operators that perform various mathematical computations (e.g., \textit{Conv}, \textit{Relu}), and some operators have attributes (e.g., stride in \textit{Conv}). Subsequent operators take the outputs of previous operators as their inputs, and the connection topology of these different operators constructs the model's computational graph. The inputs and outputs of operators are four-dimensional tensors in $[batch\_num,channel,height,width]$ format.
During model inference, data is propagated through the computational graph until the final output is obtained.
Quantization is a technique that reduces the precision of model weights. This process significantly decreases memory and computational requirements, enabling efficient inference on resource-limited edge devices, typically with minimal loss in accuracy.

DL compilers provide valuable solutions for deploying DNNs on edge devices. DL compilers perform a series of optimizations on these operators and ultimately compile them into executables. The operators are compiled into functions in executables, the inputs and ouputs of a operator correspond in sequence to the parameters of the operator function. The compiled executables can efficiently perform inference on resource-constrained edge devices. DL compilers also support quantized compiled models, which further reduces the resource consumption of model inference. The general workflow of a DL compiler involves three steps:
\emph{frontend processing}, which converts general model representations, such as ONNX~\cite{ONNX2025}, into  computational graphs supported by the compiler's frontend; \emph{compilation optimization}, applying various optimization techniques, including high-level optimizations like operator fusion and constant folding, and low-level optimizations such as layout rearrangement; \emph{code generation}, which generates executables adapted for the target device's hardware. 

\subsection{DNN Executables Decompiler}
The goal of DNN decompiler is to reverse DNN executables into identical high-level models. The typical workflow of DNN executables decompiler involves the three steps: \emph{operator type recognition}, determining the specific type of DNN operator functions; \emph{operator attribute recovery}, which recovers attribute values for operators, such as the stride for \textit{Conv}; \emph{model reconstruction}, rebuilding the computational graph and collecting operator weights to reconstruct high-level models within a general DL framework like PyTorch~\cite{paszke2019pytorch}.

\subsection{Research Gap}
DNN executables are typically deployed offline on edge devices, and attackers can reverse them to high-level DNN models, which can damage the intellectual property of the model owners, especially for models trained on private data\cite{nayan2024sok}. Moreover, based on the recovered high-level models, attackers can employ methods such as white-box adversarial attacks\cite{goodfellow2014explaining,moosavi2016deepfool,liu2023gradient} and backdoor attacks\cite{salem2022dynamic,li2022backdoor} on the DNN executables. These potential attacks can be directly implemented by editing executables and are difficult to detect, further increasing security risks.

Previous works~\cite{zhang2023libsteal,wu2022dnd,liu2023decompiling,shi2024research,wuneuroscope} have attempted to decompile DNN binaries. However, each of these methods has its limitations. We summarize the previous works and compare them with \toolname{} in Table \ref{tab:previous-work}.
\begin{table}[t]
    \centering
    \caption{Comparison with Existing DNN Decompilers}
    \label{tab:previous-work}
    \begin{tabular}{ccccc} \toprule
         \textbf{Works} & \textbf{Optimization} & \textbf{Cross Arch} & \textbf{Quantization}  \\ \midrule
         Libsteal~\cite{zhang2023libsteal} & \Circle & \Circle & \Circle  \\
         Shi et al~\cite{shi2024research}& \Circle & \Circle  & \Circle \\
         DND~\cite{wu2022dnd} & \Circle & \CIRCLE & \Circle  \\
         Neuroscope~\cite{wuneuroscope} & \Circle & \CIRCLE  & \Circle   \\
         BTD~\cite{liu2023decompiling} & \CIRCLE & \Circle & \Circle   \\
         \toolname{} & \CIRCLE & \CIRCLE & \CIRCLE   \\
         \bottomrule
    \end{tabular}
\end{table}
Libsteal cannot decompile standalone DNN executables and is unable to handle compilation optimizations. The accuracy of the models recovered by Libsteal is relatively low. The work by Shi et al. only supports x86 architecture and cannot recover models with high accuracy. DND and Neuroscope do not effectively handle compilation optimizations. Although BTD considers the impact of compilation optimizations, it only supports x86 architecture. Moreover, all previous works overlook models compiled with quantization, which limits the applicability of these methods in practical scenarios. Previous DNN executables decompilers have their own limitations and existing decompilers struggle to \textbf{simultaneously address compilation optimizations, support different architectures, and accommodate quantized compiled models.} 

Beyond the aforementioned discussion, we conduct a systematic analysis of operator type recognition, the critical step in  decompilation pipeline. This analysis results in the following observations:

\textbf{Observation1:} Most previous works lack robust support for operator type recognition. Libsteal and Shi et al.'s work do not guarantee sufficient accuracy; DND relies on symbolic execution, which incurs significant overhead and limits the size of supported models; Neuroscope only supports 12 DNN operators. More importantly, these methods fail to provide adequate support for fused operators from compilation optimizations.

\begin{figure}[t]
    \centering
    \includegraphics[width=\linewidth]{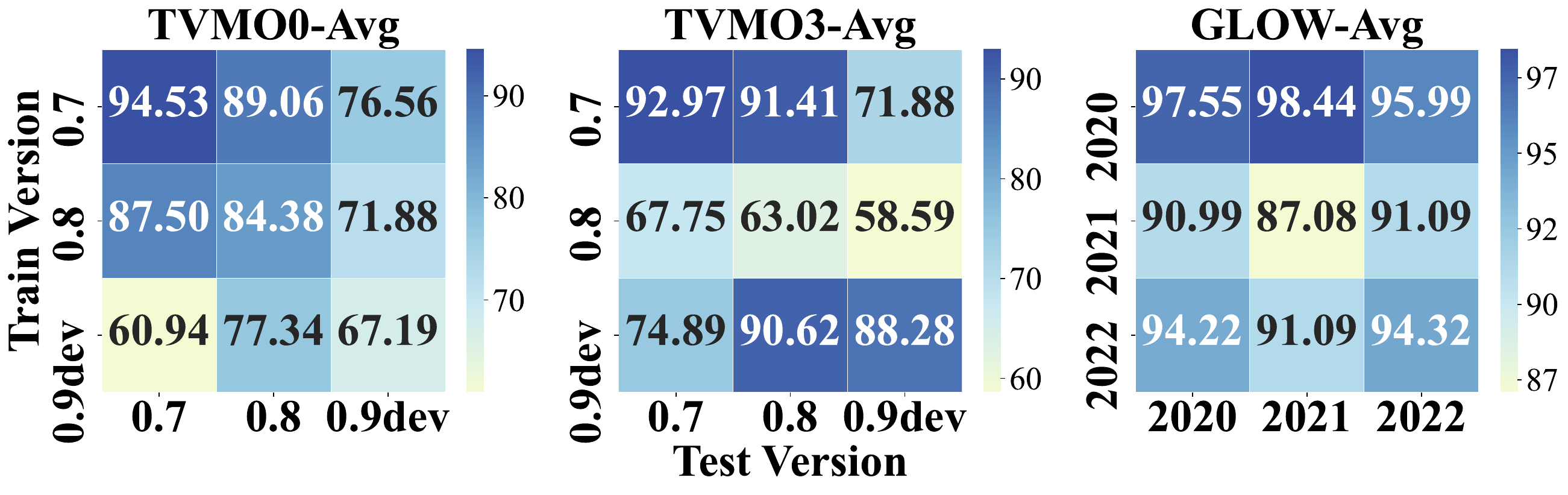}
    \caption{BTD Cross Version Type Recognition Accuracy}
    \label{fig:btdresult}
\end{figure}

\textbf{Observation2:} BTD considers the impact of compilation optimizations and supports a wider range of operator types. However, BTD trains machine learning model for each compiler version to make predictions. This approach relies heavily on training data and treats the compiler version as prior knowledge, which limits its scalability in real world scenario. Moreover, we find that about 57.9\% of the training data in TVM and 18.3\% in GLOW appear in the test dataset, further undermining the effectiveness of the method. To avoid the influence of data leak, we randomly split all the data with an 8:2 ratio for train dataset and test dataset, and retrain the model for each compiler version, strictly following
the default settings of BTD. As illustrated in Figure \ref{fig:btdresult}, we have analyzed BTD’s type recognition accuracy across different compiler versions. It is evident that BTD’s method exhibits poor cross-version support for compilers after avoiding data leak. Even within the same version, errors of operator type occur to an unacceptable degree, make it difficult to apply directly.

\begin{figure}[t]
    \centering
    \includegraphics[width=\linewidth]{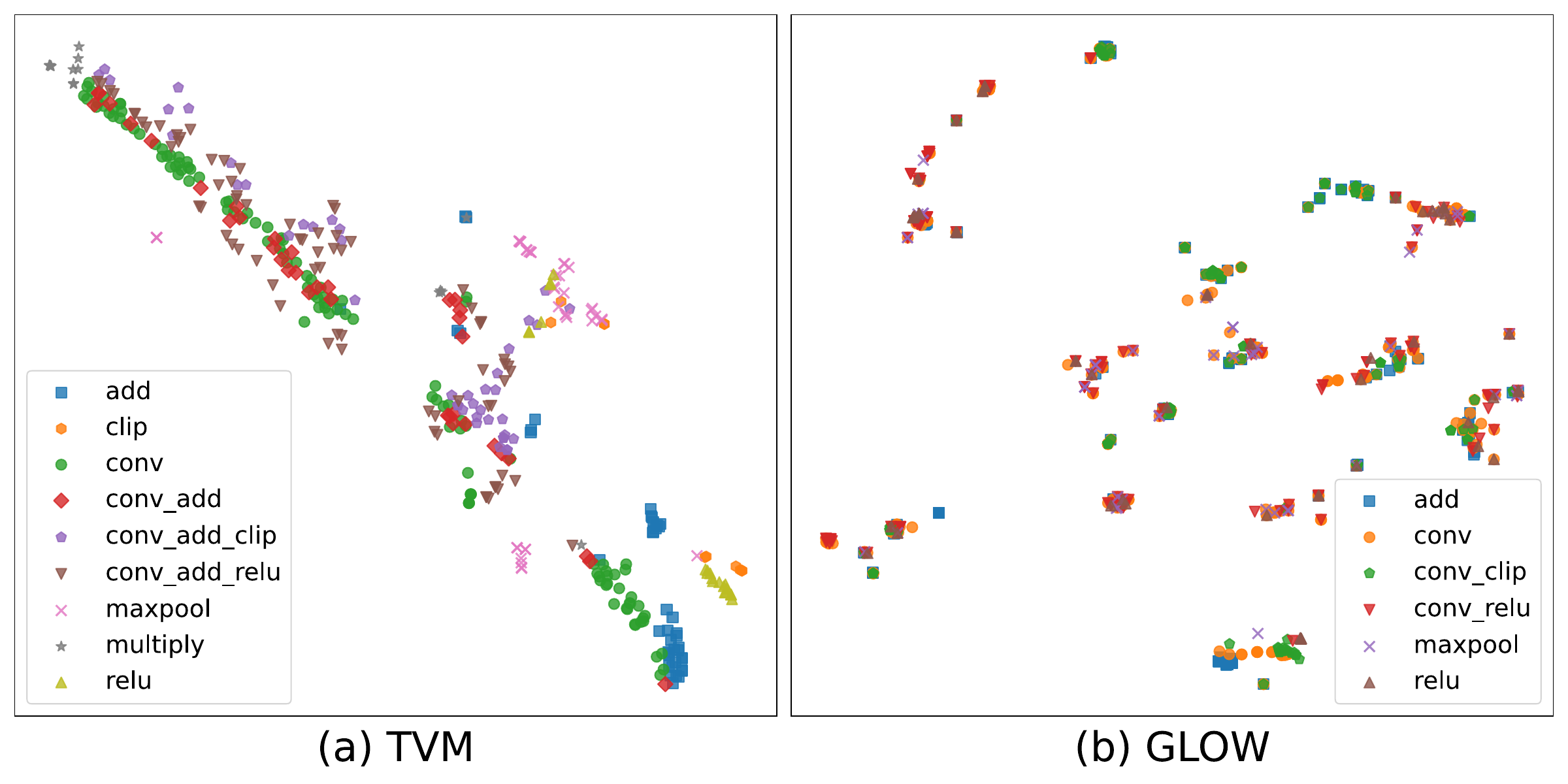}
    \caption{TSNE Dimensionality Reduction of Different Operators}
    \label{fig:tsne}
\end{figure}

\textbf{Observation3:} To overcome the reliance on prior knowledge and improve recognition accuracy, we attempt to use a mathematical feature-based approach for operator type recognition. We specifically calculate the proportion of arithmetic operators (e.g., +, -, *, /) within the decompiled code of operator functions, and apply TSNE dimensionality reduction. The result is shown in Figure \ref{fig:tsne}, the distributions of different types significantly overlap, indicating that classifying operators solely based on their mathematical characteristics is evidently challenging. Successfully performing operator type recognition requires a thorough capture of the semantic features of operator functions. The operator type recognition mechanism should have a good grasp of code semantic rather than strict mathematical feature matching.

Reflecting on these observations, \textbf{existing works struggle with operator type recognition. They exhibit limitations in the scope of covered operators and recognition accuracy.}

We summarize the challenges faced as follows. 
\textbf{C1:} A method for accurate operator type recognition needs to be designed. It should be able to cover a wide variety of operators and fused operators under compilation optimization. The method should not rely on prior knowledge such as compiler version or training data. \textbf{C2:} A decompiler needs to integrate the new method for operator type recognition, forming an end-to-end pipeline that recover DNN executables into high-level models. The decompiler should also have cross-architecture support capabilities. \textbf{C3:} Quantized compiled models exhibit new characteristics different from standard models, involving quantization scaling factors between integer and float domains. The decompiler needs to be compatible with these differences.

To address these challenges, we design \toolname{} to implement a more comprehensive decompiler for DNN executables.
\textbf{For C1:} We systematically analyze the characteristics of operators in DL compilers and design a progressive operator type recognition strategy. \toolname leverages dynamic analysis and code semantic understanding from LLMs to support compatibility with various types of operators and fused operators. \textbf{For C2:} Based on the operator type recognition method, we subsequently implement operator attribute recovery and model reconstruction, forming a complete decompilation pipeline compatibility with various types of different models. The core technology of \toolname{} is hardware-platform independent, ensuring its cross-architecture support capabilities. \textbf{For C3:} \toolname specifically adapts the model reconstruction method to convert integer domain weights back to float domain. \toolname employs a learning-based weights recovery approach that requires only a small amount of training data to recover functionally very similar models.

\subsection{Threat Model}
\toolname{} is designed towards DNN executables deployed on edge devices, where \toolname{} can access DNN executables compiled by DL compilers and extract the complete executables. DNN executables are generated through standard DL compiler pipeline with optional compiler optimization. \toolname{} has the capability to execute the executables and monitor memory status during execution. \toolname{} requires no prior knowledge of model architecture or weights, it only needs trivial inputs that satisfy the expected input format. The ultimate goal of \toolname{} is to decompile executables into identical white-box high-level DL models, effectively extracting the computational graph, weights and other information.

The threat model used in our study is consistent with previous works~\cite{shi2024research,liu2023decompiling,wu2022dnd,wuneuroscope} and is generally common and practical in real-world scenarios. The design of \toolname{} aims to highlight security risks of DNN executables and promote the safe use of DL compilers.

\section{Approach}

\begin{figure}[t]
    \centering
    \includegraphics[width=\linewidth]{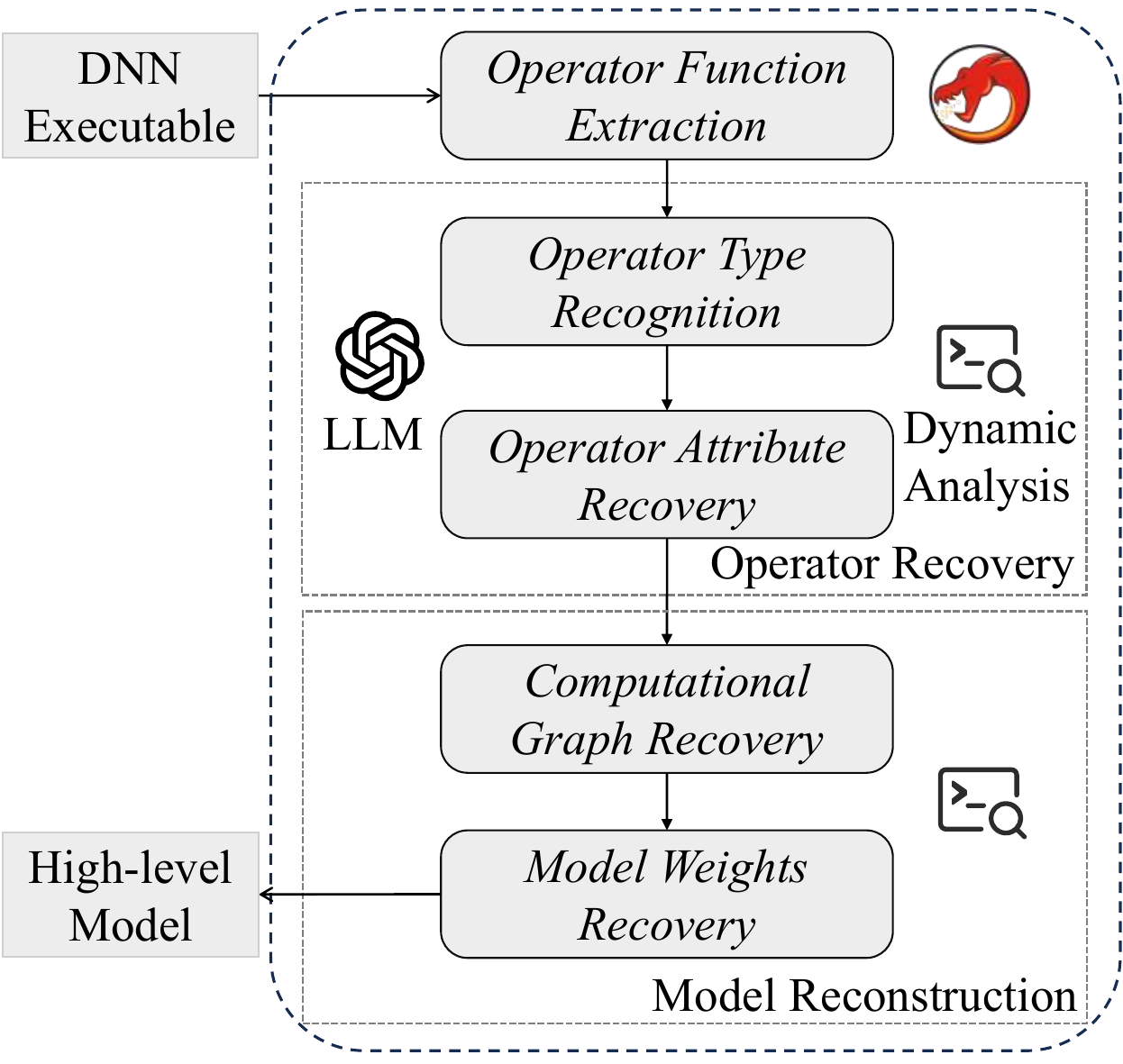}    
    \caption{Workflow of \toolname{} }
    \label{fig:overview}
\end{figure}

As shown in Figure~\ref{fig:overview}, \toolname{} is designed as an universal pipeline for different types of DNN executables, enabling end-to-end recovery of DNN executables into high-level models. Specifically, \toolname{} first extracts operator function information, such as disassembled code, decompiled code and parameter dimensions from DNN executables. In this stage, \toolname{} utilizes Ghidra~\cite{ghidra2025}, a general-purpose decompiler. 
Secondly, \toolname{} completes operator type recognition and operator attribute recovery, where it leverages dynamic analysis and code semantic understanding from LLMs to support compatibility with various types of models. Dynamic analysis aims to monitor the runtime information of operator functions. Dynamic analysis in \toolname{} requires only trivial input that satisfies the expected input format. This is due to the fact that any input can guarantee full coverage of the whole DNN model, and the mathematical dependencies of intermediate features are fixed. LLMs can understand the mathematical semantics of different operators in decompiled code without relying on prior knowledge like compiler versions or training data. Finally, \toolname{} performs model reconstruction by computational graph and model weights recovery, recovering high-level models. Next, we will sequentially introduce the components of \toolname{}.

\subsection{Operator Function Extraction}
\toolname{} first collects operator function information  including disassembled and decompiled code from DNN executables using ghidra.

\begin{figure}[t]
    \centering
    \includegraphics[scale=0.5]{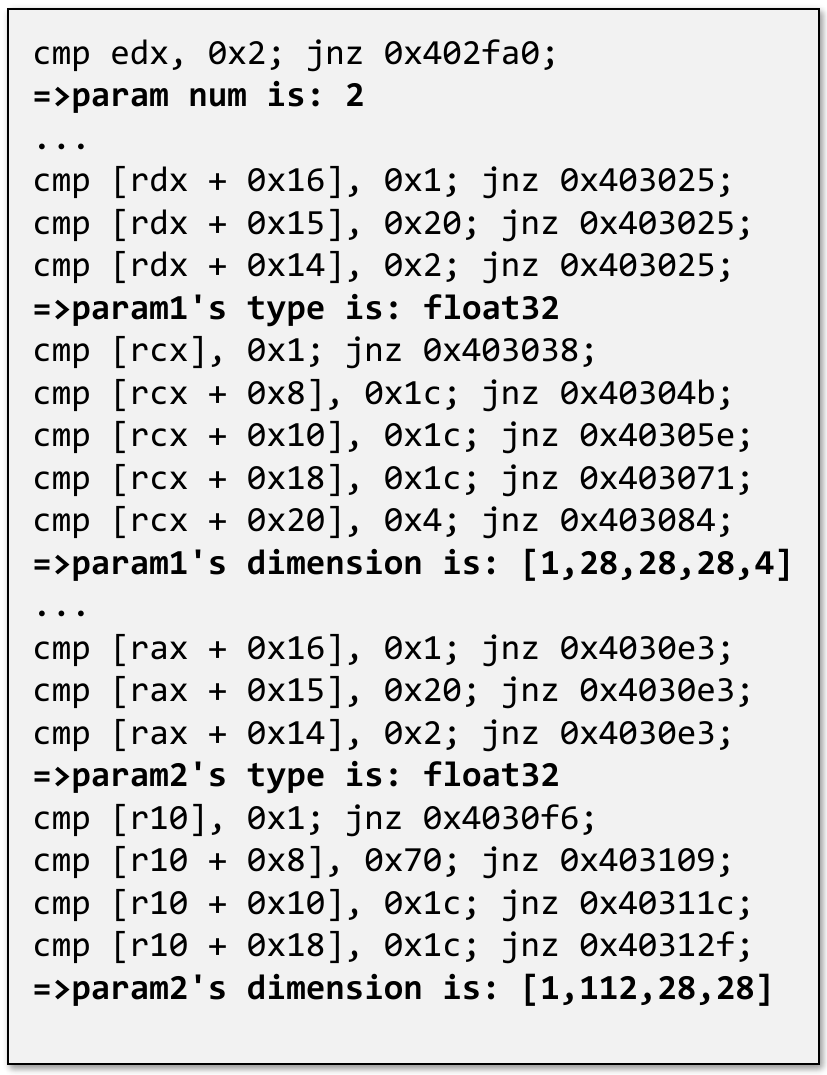} 
    \caption{An Example of Disassembled Code}
    \label{fig:example}
\end{figure}

Inspired by previous works~\cite{zhang2023libsteal,shi2024research}, \toolname{} can identify the dimensions of operator parameters from disassembled code in TVM compiler. \toolname{} further expands on their methods, \toolname{} also extracts the types of operator parameters and recover the optimized parameters' dimensions. The parameters' types can help determine the weights of quantized compiled models. The recovery of optimized parameters' dimensions can help analyze optimized operators. Figure~\ref{fig:example} is a specific example of \textit{Transform} operator. The parameters' dimensions and types can be extracted by scanning the disassembled code.

The dimension of the first parameter is optimized into $[N,C,H,W,c]$ layout. NCHWc~\cite{chen2018tvm} is a commonly used inference optimization method, which adapts to hardware and accelerates inference. \toolname{} can recover it into standard $[N,C*c,H,W]$ format. Similarly, the dimension $[O_c, I_c, K, K]$ of a \textit{Conv} kernel parameter may be optimized into $[O_c/A, I_c/B, K, K, B, A]$~\cite{chen2018tvm}. The optimized layout decides the storage order of model weights in memory. \toolname{} extracts this optimized dimension layout and converts it to standard format. This method is layout agnostic, it is compatible with other optimized layout, such as $[N,H,W,C]$. The parameters of the function correspond sequentially to the inputs and outputs of the operator. Each type operator has a fixed number of inputs and outputs, and \toolname{} can accurately recover parameters' dimensions. We validate this characteristic across both historical and recent TVM versions, ensuring the generality and robustness of our approach.

\subsection{Operator Type Recognition}
The purpose of operator type recognition is to determine the specific type of the operator function. 

We manually analyze DL compiler's support for DNN operators~\cite{chen2018tvm,rotem2018glow} and align it with the practice of general DL frameworks like ONNX~\cite{ONNX2025} to classify operators. The operators can be divided into four types as shown in Table~\ref{tab:operators}. Type 1 is \textit{Layout Transformation}. This type of operator adjusts the layout of input tensors and it can be further divided into inter-tensor types like \textit{Concat} and intra-tensor types like \textit{Flatten}. Inter-tensor types involve layout adjustments across multiple tensors, whereas intra-tensor refers to layout adjustments only to a single input tensor. Type 2 is \textit{Element-wise}. This type of operator performs element-wise computation on the input tensor. It can be further divided into type 2.1 with two operands (one input and one output) like \textit{Softmax} and type 2.2 with three operands (two inputs and one output) like \textit{Add}. Type 3 is \textit{Reduction}. This type of operator performs feature reduction on the input tensor, such as \textit{Maxpool}. The output tensor is usually compressed in size compared to the input tensor. Type 4 is \textit{Complex}. This type of operator performs complex and intensive computations, such as \textit{Conv}. Due to the optimization of DL compilers, multiple operators may be fused into one operator to accelerate inference. TVM refers to enabling no optimization as ``O0", while enabling full optimizations as ``O3". GLOW applies full optimizations by default. Next, we will introduce the specific methods of operator type recognition towards TVM and GLOW.

\begin{table}[t]
\centering
\caption{Operator Classification}
\label{tab:operators}
\begin{tabular}{m{1.5cm} m{2.5cm}  m{3.5cm}} 
\toprule
\multicolumn{1}{c}{\textbf{Type}} & \multicolumn{1}{c}{\textbf{Operators}} & \multicolumn{1}{c}{\textbf{Description}}\\ 
\midrule
1.Layout \newline Transformation & 1.1 concat,split... \newline 1.2 flatten,reshape,\newline transpose... & Inter-tensor layout transformation\newline Intra-tensor layout transformation\\
\midrule
2.Element-wise          & 2.1 softmax,sqrt,clip,\newline abs,relu... \newline 2.2 add,sub,mul,div,\newline power... & Element-wise computation of two operands \newline Element-wise computation of three operands\\
\midrule
3.Reduction             & 3.1 maxpool,avgpool,\newline sum,max... & Feature reduction and extraction \\
\midrule
4.Complex               & 4.1 conv,dense... & Complex and intensive computation \\
\bottomrule
\end{tabular}
\end{table}

For TVM, as parameters' dimensions of operator function can be collected during the operator function extraction phase, \toolname{} uses them to enhance operator type recognition. Operators can undergo coarse-grained identification of their types. For \textit{Layout Transformation} type and \textit{Complex} type, specific operator types can be directly determined. For example, an operator with two parameters' dimensions: $[1, 1000, 1, 1]$ and $[1, 1000]$ can be directly identified as \textit{Flatten}. For \textit{Conv}, input channel and output channel of kernel parameter correspond respectively to the channel of the first and last parameters. For \textit{Element-wise} and \textit{Reduction} type, operators can be determined candidate list among 2.1, 2.2, 3.1 in Table~\ref{tab:operators} according to the number and dimension of parameters. Further, leveraging the code semantic understanding capabilities of LLMs, \toolname{} determines the specific operator types within the candidate list based on the mathematical features of the decompiled code. Operator type recognition needs code semantic understanding rather than strict mathematical feature matching, LLMs are more suitable for it. This approach overcomes the limitations of previous methods, which rely heavily on training data and struggle with the cross-version support for compilers. \toolname{} does not rely on complex prompt design, a simple and clear prompt can effectively accomplish the task, ensuring the stability of LLM participation in \toolname{}. Here is the simplified prompt for specific operator type recognition:

\begin{tcolorbox}[
  colback=white!95!black, 
  colframe=white!10!black, 
  title=\textbf{Simplified prompt of operator type recognition}, 
  fonttitle=\bfseries,
  halign title=center,
  rounded corners,
  boxrule=1pt,
]

You are an AI assistant specialized in analyzing operators.
Given the decompiled code of an operator: \{pseudocode\}, identify the operator as one of \{candidates\}.
\end{tcolorbox}
At the O3 optimization level, operators frequently undergo fusion. \toolname{} identifies their types through dynamic analysis.
The most common operator fusion occurs \textit{Conv} and \textit{Dense} with the operators following them. We use them as an example to illustrate how \toolname{} handles fused operators.  As shown in Algorithm \ref{alg:fuse}, for \textit{Conv} and \textit{Dense}, during operator fusion, they are fused with several \textit{Multiply} or \textit{Add} operators, which are typically related to bias addition, batch normalization, etc. Specifically, \toolname{} records the parameter addresses of each operator function through dynamic instrumentation. If the input is the output of previous operators, this indicates a skip connection (jumpadd). Otherwise, \toolname{} starts recording the memory access during the operator function execution, identifying the instruction that initially accesses the parameter address. From this instruction, \toolname{} performs taint analysis, tracking relevant registers until first encounter a multiply or add instruction. Activation functions like \textit{Relu} and \textit{Clip} are often attached to the tail of the fused operator with repeated patterns in decompiled code. \toolname{} extracts the tail part of the decompiled code and uses LLM to verify if activation is accompanied by the fused operator. \toolname{} employs the same processing method for other fused operators like the operator consisting of \textit{Concat} and activation. \toolname{} provides more comprehensive analysis and support for fused operators than previous works.

\begin{algorithm}[t]
\caption{Fused Operator Type Recognition}
\label{alg:fuse}
\KwIn{Operator: $op (conv/dense)$}
\KwOut{Operator\_type: $fuse\_type$}
\SetKwComment{Comment}{$\triangleright$\ }{}
$fuse\_type \leftarrow op.base\_type$\;
$param\_addrs \leftarrow \text{log\_param\_addrs}(op)$\Comment*[r]{ record function parameter addresses}

\For{$i = 2$ \KwTo $param\_addrs.\text{length} - 1$}{
    
    \If{$param\_addrs[i] \in \text{prev\_ops.output}$}{
        $fuse\_type \leftarrow fuse\_type + $ ``jumpadd''\;
    }
    \Else{
        $start \leftarrow \text{mem\_read\_first}(op, param\_addrs[i])$\Comment*[r]{record the instruction address that first reads the param}
        $fuse\_type \leftarrow fuse\_type + \text{taint}(op, start)$\Comment*[r]{ taint propagation until determine mul or add}
    }
}
$fuse\_type \leftarrow fuse\_type + \text{check\_activation}(op)$\Comment*[r]{check activation and activation\_type}
\Return{$fuse\_type$}
\end{algorithm}

For GLOW, the disassembled code does not contain dimension information, but the optimization strategy in GLOW is relatively simple. Operator fusion almost only occurs after \textit{Conv} with activation. \toolname{} directly performs operator type recognition by classifying the decompiled code with LLM, which 
considers all operators as the candidate list and determines the specific operator types within the candidate list based on the mathematical semantics of operators.

Currently, \toolname{} supports the most common operators of ONNX (including but not limited to operators analyzed in previous works), the complete list of operators is shown below. 

\begin{RQBox}
\noindent\textbf{TVM:} 
\noindent concat, split, pad, flatten, transform, reshape, expand\_dim, transpose, add, sub, mul, div, power, softmax, sqrt, rsqrt, flatten, clip, neg, abs, lrn, relu, exp, maxpool, avgpool, sum, max, conv2d, dense. 

\noindent\textbf{TVM\_fused:} 
\noindent conv2d $\cdot$ (mul\textbar add)* $\cdot$ (activation)?, dense $\cdot$ (mul\textbar add)* $\cdot$ (activation)?, add $\cdot$ (activation)?, concat$\cdot$ (reshape $\cdot$ transpose $\cdot$ reshape)? $\cdot$ (activation)?, concat $\cdot$ (reshape $\cdot$ transpose $\cdot$ reshape)? $\cdot$ split, reshape $\cdot$ transpose $\cdot$ reshape.

\noindent\textbf{GLOW:}
\noindent maxpool, avgpool, softmax, relu,  lrn, add, sub, mul, dense, conv2d, convdkkc8, conv2d\_relu, conv2d\_clip, 

\noindent tensor\_transformation (insert\_tensor, extract\_tensor...).
\end{RQBox}
We compile and decompile validated computer vision models from the ONNX Model Zoo, and \toolname{} can cover all the operators of them. The operator type recognition method of \toolname{} has good scalability. For more operators, it only requires defining their types in Table~\ref{tab:operators}. The pipeline of operator type recognition in \toolname{} is universal and scalable.
\subsection{Operator Attribute Recovery}
Some operators contain specific attribute value, such as the stride and padding of \textit{Conv}. To ensure the functional consistency of the reconstructed model, it is essential to accurately recover these attributes. \toolname{} combines dynamic analysis and code semantic understanding from LLM to design attribute recovery methods for different operators.

For \textit{Conv}, \toolname{} utilizes parameters' dimensions of these operators to recover attributes. For TVM, dimensions have been collected from disassembled code introduced in operator function extraction. For GLOW, \toolname{}
employs dynamic analysis to collect parameters' dimensions.
The \textit{Conv} in GLOW has four parameters: \( out \), \( in \), \( weight \) and \( bias \). \toolname{} records the memory read ranges starting from these four parameters. \toolname{} first extracts kernel\_size from decompiled code, the kernel\_size is manifested as two continuous outer loop with identical values in decompiled code. The output channel is equal to the number of weights read from the bias parameter. Next, the following relationships can be used: 
\begin{equation}
   \text{in\_channel} = \frac{\text{weight\_region}}{\text{out\_channel} \times \text{kernel\_size}^2} 
\end{equation}

For \textit{Maxpool} and \textit{Avgpool}, \toolname{} extracts channel value from decompiled code, which corresponds to the inner loops following two identical loops. \toolname{} records the ranges of input and output. Next, the following relationships can be used:

\begin{equation}
 \text{in\_height}/\text{out\_height} = \sqrt{\frac{\text{region}}{\text{channel}}}  
\end{equation}

\begin{algorithm}[t]
\caption{Operator Attribute Recovery}
\label{alg:attr_extract}
\KwIn{List of operators: $ops$}
\KwOut{List of attributes: $attrs$}
\SetKwComment{Comment}{$\triangleright$\ }{}

$attrs \leftarrow [\;]$\;

\ForEach{$op \in ops$}{
    \If{``maxpool'' $\in op.type$}{
        $in, out \leftarrow \text{dump\_io}(op)$\Comment*[r]{ dump input and output tensors of the op}
        
        \For{$k = 1$ \KwTo $in.width$}{
            $stride, padding \leftarrow \text{infer\_sp}(k)$\Comment*[r]{ infer stride and padding}
            
            \If{$\text{sim\_forward}(in, out, k, stride, padding)$}{
                $attrs.\text{append}([k, stride, padding])$\Comment*[r]{ simulate forward of the op and append when output match}
                \textbf{break}\;
            }
        }
    }
    \ElseIf{``avgpool'' $\in op.type$}{
        $k \leftarrow \text{extract\_attr}(op.decompiled)$\Comment*[r]{ extract attributes using LLM}
        
        $stride, padding \leftarrow \text{infer\_sp}(k)$\;
        $attrs.\text{append}([k, stride, padding])$\;
    }
    \ElseIf{``conv'' $\in op.type $ }{
        $stride, padding \leftarrow \text{infer\_sp}(k)$\;
        $attrs.\text{append}([k, stride, padding])$\;
    }
    \ElseIf{``lrn'' $\in op.type$ \textbf{or} ``clip'' $\in op.type$}{
        $attr \leftarrow \text{extract\_attr}(op.decompiled)$\;
        $attrs.\text{append}(attr)$\;
    }
    \ElseIf{``concat'' $\in op.type$ \textbf{or} ``transform'' $\in op.type$}{
        $in, out \leftarrow \text{dump\_io}(op)$\;
        $attr \leftarrow \text{sim\_forward}(in, out)$\;
        $attrs.\text{append}(attr)$\;}

}
\Return{$attrs$}
\end{algorithm}
The recovery of parameters' dimensions in GLOW is also layout agnostic like TVM. \toolname only needs the size of accessed memory region and does not concern the specific layout order.
Algorithm \ref{alg:attr_extract} explains how \toolname{} accomplishes operator attribute recovery. For \textit{Conv} or \textit{Pooling} operators, the input\_height, output\_height and kernel\_size are inherently contained within the parameters' dimensions. The input\_height \(I_{h}\), padding \(P\), \( \text{kernel\_size} \) \(K\), stride \(S\), and output\_height \( O_{\text{h}} \) satisfy the following constraint:
\begin{equation}
    O_h = \left\lfloor \frac{(I_h + 2P - K)}{S} \right\rfloor + 1
\end{equation}
Except for stride and padding, all other variables for \textit{Conv} are known, and both stride and padding must be integers. To determine the values of them, \toolname{} employs a constraint enumeration method. Starting with \( \text{stride} = 1 \) and \( \text{padding} = 0 \), \toolname{} enumerates different combinations in ascending order to find solution that satisfies the constraint.

For \textit{Avgpool} and \textit{Maxpool}, \( \text{kernel\_size} \) does not explictly appear in the dimension information. 
In the case of \textit{Maxpool}, kernel\_size is reflected in the number of max-related instructions. However, due to compilation optimizations, it is hard to accurately infer the kernel\_size directly. To precisely recover the attributes of \textit{Maxpool}, \toolname{} instruments the executable to record the input and output tensors of \textit{Maxpool}. \toolname{} then enumerates different combinations of kernel\_size, stride and padding, simulates the forward of the \textit{Maxpool} until the computed tensor exactly matches the actual output tensor. It is worth noting that any trivial input can achieve full coverage, so \toolname{} only needs one trivial input to simulate forward. In the case of \textit{Avgpool}, kernel\_size is evidently reflected in the decompiled code. For example, patterns like ``\(*0.020408(1/49)\)'' repeatedly appear, indicating that \( \text{kernel\_size} \) is 7. LLM can extract kernel\_size from decompiled code to reduce the overhead of dynamic analysis. \toolname{} infers stride and padding of \textit{Avgpool} through the constraint enumeration method same with \textit{Conv}.

Local response normalization (lrn) has attributes: \(size\), \( \beta \), \( \alpha \), \( bias \) and \textit{Clip} has attributes:  \( min \), \( max \). The attributes of lrn and clip generally have a large search space, making them unsuitable for simulation execution using dynamic analysis enumeration. However,  these attributes are often obvious in the decompiled code, which can be extracted directly by LLM. Regarding \textit{Concat}, the sequence of multiple inputs is an attribute that should be extracted. Moreover, at the O3 optimization level in TVM, \textit{Transpose} may be fused after \textit{Concat}. \toolname{} instruments the executable to record input and output tensors of \textit{Concat}. By enumerating the input order and the number of channel shuffle groups in \textit{Transpose}, \toolname{} simulates the forward of \textit{Concat} until the computed tensor exactly matches the actual output tensor. Similarly, \textit{Transform} may also involve \textit{Transpose} fused afterward, \toolname{} determines the attribute values same with \textit{Concat}.
\toolname only requires straightforward prompt for operator attribute recovery. Here is the simplified prompt in operator attribute recovery:

\begin{tcolorbox}[
  colback=white!95!black, 
  colframe=white!10!black, 
  title=\textbf{Simplified prompt of operator attribute recovery}, 
  fonttitle=\bfseries,
  halign title=center,
  rounded corners,
  boxrule=1pt
]

You are an AI assistant specialized in analyzing operators.
Given the decompiled code of a \{type\} operator: \{pseudocode\}, infer the \{attribute\}.
\end{tcolorbox}

\subsection{Model Reconstruction}
After recovering operator types and attributes, it is also necessary to restore the computational graph and weights to ultimately construct the high-level model. 

To restore the computational graph, \toolname{} instruments the executable to record the execution sequence of all operator functions and the parameter addresses at each operator's entry point. All operators have fixed calling conventions, including the number of input and output parameters. By matching the input and output addresses between different functions, We can determine the input-output connection relationships between different operators. The computational graph of the model can be effectively restored finally. For the recovery of model weights, 
\toolname{} dumps weights from weight parameter addresses and adjusts the weight layout according to their dimension information. Finally, \toolname{} converts recovery operators into the corresponding PyTorch model, resulting in a identical reusable high-level model.

During model reconstruction, \toolname needs to adapt to the specific characteristics of quantized compiled models. Quantized compiled methods involves global\_scale mode and kl\_divergence mode. Scale refers to the scaling factor between the float domain and int domain during quantized compilation. The global\_scale method uses the same scale across different operators, and does not rely on data calibration. The kl\_divergence method needs data for calibration, resulting in higher precision. The weights of quantized compiled models are in the integer domain, and model inference is in integer domain as well. \toolname{} makes certain adaption to model weights recovery, \toolname{} also records the parameter types and transforms dumped bytes into corresponding type. Our goal is to recover a reusable, general model with float weights. During model inference, we observe that weights in the integer domain are updated through shift multiplication (e.g., \({weight * (0x6f65c500 >> 0x28)}\)) to prevent integer overflow. \toolname{} extracts shifted multiplication value from decompiled code, and compute updated weights: \({updated}_{w} = {dumped}_{w} \times {shifted\_multiplication}\).
 
\begin{figure}[t]
    \centering
    \includegraphics[width=\linewidth]{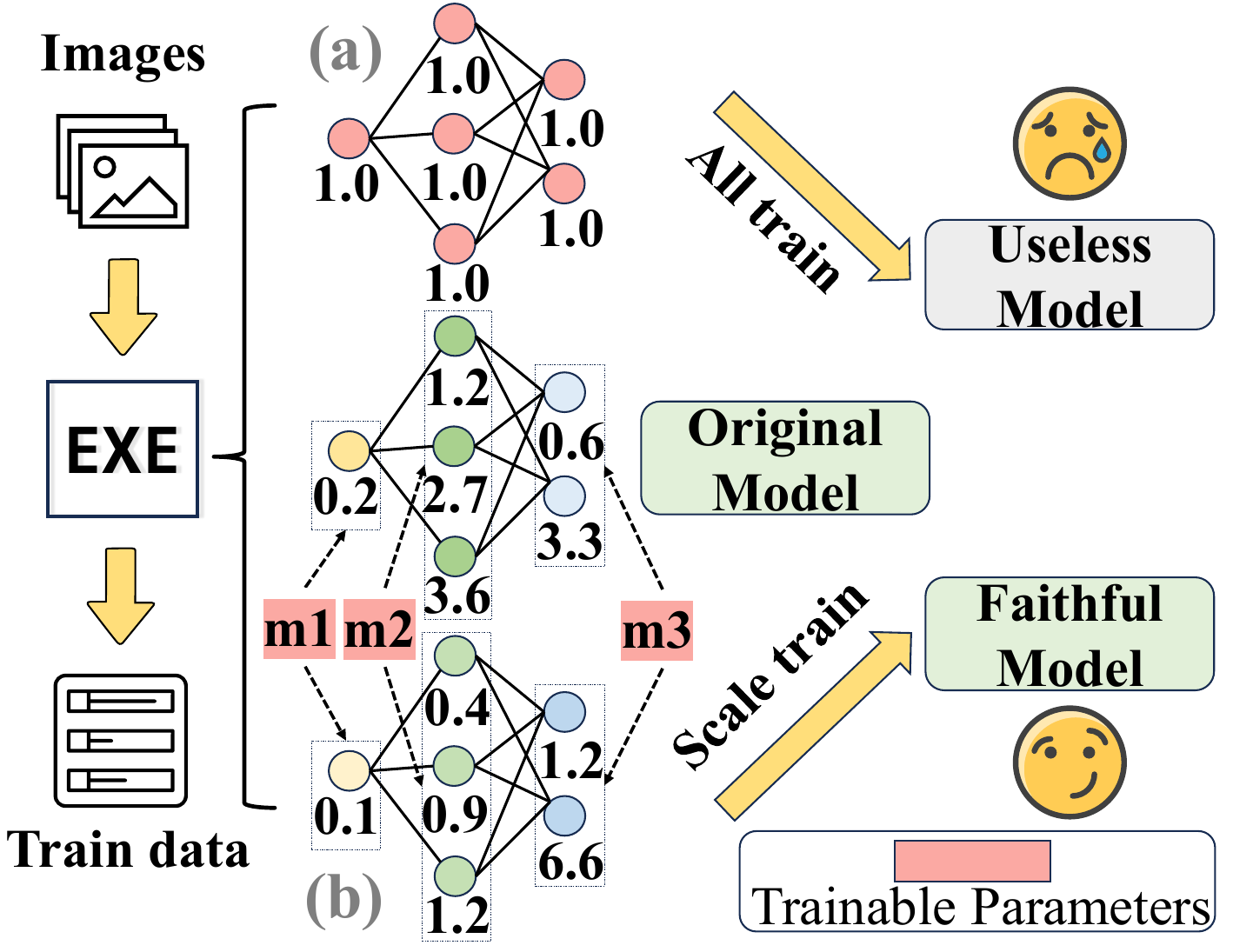}
    \caption{Analysis of Quantized Compiled Models }
    \label{fig:quantmethod}
\end{figure}

For the global\_scale method, different operators have same quant\_scale, so the updated weights are close to the original model's weights. For the kl\_divergence method, quant\_scale adjusts for different operators. Due to compilation optimizations, different quant\_scale can be confused with regular multiplications, making it difficult to identify them. Eventually, the updated weights are a multiple of the actual weights, and different layers have different multiples. To address this, \toolname{} adopts a substitute training method illustrated in Figure~\ref{fig:quantmethod}. The target model is the quantized compiled DNN executable, \toolname{} first prepares query dataset to collect labels in the target executable. Next, \toolname{} freezes known weights and trains the multiples between updated weights and actual weights. Typical black-box substitute method for model extraction (a in Figure~\ref{fig:quantmethod}) requires retraining all the model weights, whereas \toolname{} only needs to train a limited multiples (b in Figure~\ref{fig:quantmethod}). A small amount of training data is enough to gain a model that effectively replicates the functionality of the target model.

\section{Evaluation Setup}
In the evaluation, we intend to answer the following questions:

\begin{enumerate}[label=\textbf{RQ\arabic{enumi}}, leftmargin=*, align=left]
    \item \textbf{(Correctness):} How does \toolname's performance in correctness compare to State-of-the-Art (SOTA) method? 
    \item \textbf{(Efficiency):} What is the overhead of \toolname{} compared to SOTA method?
    \item \textbf{(Comprehensiveness):} Can \toolname{} cover a wider range of models and different architectures?  
    \item \textbf{(Robustness):} Can \toolname{} fix errors encountered during the decompile process and what is the impact of different LLMs on \toolname{}?

\end{enumerate}
\subsection{Implementation \& Environment}

We implement \toolname{} with about 8K LOC Python code and about 1K LOC C++ code.
In our experiments, we select GPT-4o~\cite{gpt-4o} for its superior performance in code understanding~\cite{zhao2024codejudge,yu2024humaneval} and the large size (128K) of context window, which is sufficient to cover the operator functions. We conduct experiments on both x86 and aarch64 architectures, we use Intel Pin~\cite{luk2005pin} for dynamic analysis in the x86 architecture, and GDB~\cite{GDB2024} in the aarch64 architecture. Both Pin and GDB are automated through Pintool scripts and GDB scripts to finish tasks like memory dump.
\toolname{} employs Ghidra~\cite{ghidra2025}, a well-known decompiler in reverse engineering, and the version is 11.1.2. \toolname{} reconstructs DNN executables into high-level models using PyTorch.
We statistics and compare the overhead of different methods on AMD EPYC 9654 CPU with 60GB RAM.

\subsection{Baseline \& Model Selection}
Libsteal and Shi et al.'s work struggle to accurately recover the model, and there is no official code implementation available for these methods. DND and Neuroscope fail to analyze executables with compilation optimizations. DND only analyzes on three small models: MobileNetv2~\cite{sandler2018mobilenetv2}, ResNetv1~\cite{he2016deep}, and MNIST~\cite{CortesLeCunBurgesMNIST}. The symbolic execution methods in DND introduces significant overhead, limiting its capability to analyze larger models. Neuroscope only supports 12 DNN operators, which limits its usability. What's more, none of these four decompilers analyze the influence of compiler versions. BTD takes into account compilation optimizations and has been tested on a larger range of models. BTD conducts extensive experiments across different optimization levels and different compiler versions, which demonstrates its effectiveness. Overall, BTD is currently the SOTA method available. 

\begin{table}[t]
    \centering
    \caption{Details for the Evaluation Models}
    \setlength{\tabcolsep}{6.5pt}
    \label{tab:models}
    \begin{tabular}{ccccc}
        \toprule
        \textbf{Model} & \textbf{Nodes} & \textbf{Edges} & \textbf{Params} & \textbf{Outs} \\ \midrule
        ShuffleNetv2~\cite{ma2018shufflenet} & 229 & 245 & 2,294,784 &1000 \\
        MobileNetv2~\cite{sandler2018mobilenetv2} & 104 & 115 & 3,487,816&1000 \\ 
        EfficientNet~\cite{tan2019efficientnet} & 188 & 215 & 12,966,032&1000 \\ 
        Inceptionv1~\cite{szegedy2015going} & 143 & 170 & 6,998,552&1000 \\ 
        ResNet18~\cite{he2016deep} & 69 & 77 & 11,699,112&1000 \\ 
        VGG16~\cite{simonyan2014very} & 41 & 41 & 138,357,544&1000\\ 
        ResNet34~\cite{he2016deep} & 125 & 141 & 21,814,696&1000 \\ 
        SqueezeNet~\cite{iandola2016squeezenet} & 65 & 73 & 1,235,496&1000 \\ 
        MnasNet~\cite{tan2019mnasnet} & 151 & 161 & 3,200,016&1000 \\ 
        ShuffleNetv1~\cite{zhang2018shufflenet} & 203 & 219 & 1,420,176&1000 \\ 
        Emotion~\cite{barsoum2016training} & 46 & 47 & 8,757,704&8 \\ 
        SuperRes~\cite{shi2016real} & 10 & 10 & 59,657 &451,584\\ \bottomrule
    \end{tabular}%
\end{table}

As shown in Table~\ref{tab:models}, to ease of comparison, we evaluate \toolname{} on six different DL models, comprising a total of 54 DNN executables (varying in compiler, optimization level, and compiler version) that are analyzed in BTD. Moreover, we have supplemented our evaluation with six more different models, aiming to cover more model structures and varying hyperparameters for same model structure. Furthermore, we conduct an additional analysis on 42 BTD unexplored DNN executables, covering a wider range of compiler versions, architectures, and quantized models. All the executables in our evaluation are stripped.

\subsection{Evaluation Metrics}
We evaluate the effectiveness of \toolname{} from three perspectives: operator type recognition, operator attribute recovery and recovered model inference.

For operator type recognition, we measure Type Recognition Accuracy (TRA). Meanwhile, for operator attribute recovery, we measure Attribute Recovery Accuracy (ARA).

Regarding model inference, we adopt the same method from previous studies~\cite{wu2022dnd,liu2023decompiling}, which compares the inference results and confidence scores of the recovered model with those of the executables. We regard the inference results as consistent only if the labels and confidence scores are exactly identical or differ solely by negligible precision loss. We employ 100 inputs for the Emotion and SuperRes models and 500 inputs for all other models, calculating Model Inference Accuracy (MIA).
For quantized compiled models, the unavoidable precision loss between integer and float domains during quantization and dequantization makes it impossible to recover a model with identical confidence scores. Therefore, we consider the inference correct if the labels match and calculate top-1, top-3, and top-5 of MIA.

\section{Evaluation Result}
In this section, we show the experimental results to answer the research questions. 
\subsection{RQ1: Correctness}

To answer RQ1, we compare \toolname{} with the SOTA method BTD, using the models involved in BTD experiments to demonstrate the performance of \toolname{}.

\begin{table*}[t]
\centering
\caption{Comparison of TRA between BTD and \toolname{} (all value in \%)}
\label{tab:comparewithbtd}
\setlength{\tabcolsep}{4.0pt}
\begin{tabular}{c *{6}{cc} c }
\toprule
\multirow{2}{*}{\raisebox{-0.5\height}{\textbf{Model}}} & 
\multicolumn{2}{c}{\textbf{EfficientNet}} & 
\multicolumn{2}{c}{\textbf{Inceptionv1}} &
\multicolumn{2}{c}{\textbf{MobileNetv2}} &
\multicolumn{2}{c}{\textbf{ResNet18}} &
\multicolumn{2}{c}{\textbf{ShuffleNetv2}} &
\multicolumn{2}{c}{\textbf{VGG16}} 
\\
\cmidrule(lr){2-3} \cmidrule(lr){4-5} \cmidrule(lr){6-7} \cmidrule(lr){8-9} \cmidrule(lr){10-11} \cmidrule(lr){12-13}
& BTD & \toolname{} & BTD & \toolname{} & BTD & \toolname{} & BTD & \toolname{} & BTD & \toolname{} & BTD & \toolname{}\\
\midrule
TVM\_v0.7\_O0

& 72.48 & 97.96 & 97.1 & 100& 80.47 &100&76.56 &100&70.05 &98.87&84.85 &100  \\ 
TVM\_v0.8\_O0 

& 39.91 & 99.28 & 87.85& 100 & 66.85 &100&46.88 &100&49.83 &98.97& 63.64&100\\ 
TVM\_v0.9dev\_O0

& 40.7 & 100 & 86.06 & 100 & 64.32 & 100& 34.92&100&39.06 &100& 53.12&100  \\ 
TVM\_v0.7\_O3

& 35.14 & 97.3 & 97.66 & 98.73 & 42.86 & 100& 78.79&96.97& 77.78&94.44&59.26 &88.9 \\ 
TVM\_v0.8\_O3

& 5 & 90 & 79.77 & 92.68 & 11.76 & 100& 54.55&100&77.14 &97.14&70.37 &100  \\ 
TVM\_v0.9dev\_O3

& 5.41& 100 & 79.22 & 100 & 2.94 & 100& 57.58&100& 71.43&100& 55.56&100  \\ 

GLOW\_2020

& 54.24& 96.61 & 77.19 & 99.04 &  70.45&95.45&57.14 &96.43&86.27 &96.08& 68.18&86.36   \\ 
GLOW\_2021

& 58.62 & 96.55 & 87.18& 98.06& 76.74& 95.35&54.29 &100& 94&100&90 &95\\ 
GLOW\_2022

& 58.62 & 96.55 & 87.18 & 100 & 76.74 & 97.67& 54.29&100& 92&100&90 &95 \\ 
\bottomrule
\end{tabular}
\end{table*}
\begin{table}[htbp]
\centering
\caption{TRA and THA of BTD (consistent with  Table~\ref{tab:comparewithbtd}) on Different Compiler Versions (all value in \%)}
\label{tab:TRATHA}
\setlength{\tabcolsep}{1.6pt}
\begin{tabular}{c *{5}{cc}  }
\toprule
\multirow{2}{*}{\raisebox{-0.5\height}{\textbf{Metric}}} & 
\multicolumn{3}{c}{\textbf{TVM\_O0}} & 
\multicolumn{3}{c}{\textbf{TVM\_O3}} &
\multicolumn{3}{c}{\textbf{GLOW}} &

\\
\cmidrule(lr){2-4} \cmidrule(lr){5-7}\cmidrule(lr){8-10}  
& v0.7 & v0.8 & v0.9dev & v0.7 & v0.8 & v0.9dev & 2020 & 2021 & 2022\\
\midrule
TRA\_Avg

& 80.4 & 64.43 & 61.31 & 70.33& 57.67 &54.86&72.57 &79.36&78.99 \\ 
THA\_Avg

& 96.42 & 94.57 & 98.50& 98.04 & 97.05 &97.10&97.33 &98.48&96.91 \\
\bottomrule
\end{tabular}
\end{table}

We remove the training data that is also on test dataset (operators of the six models in Table~\ref{tab:comparewithbtd}) and retrain the BTD operator type recognition models following the default settings. The comparison results of TRA are shown in Table~\ref{tab:comparewithbtd}. The TRA of \toolname\ significantly outperforms BTD, providing a crucial foundation for accurate operator type recognition, which is essential for subsequent operator attribute recovery and model reconstruction. \toolname{} achieves results very close to 100\% across all experimental models, whereas the BTD method faces relatively severe errors.

As a multi-classification machine learning model, BTD can also be evaluated based Type Hamming Accuracy (THA): \(\frac{1}{N} \sum_{i=1}^{N} \left( \text{Pred}_i == \text{Label}_i \right) \)
, where \( N \) denotes the number of operator classes, and it's also the length of the prediction vector for a single sample. The average result of TRA and THA 
for different compiler versions are shown in Table~\ref{tab:TRATHA}. Although THA may achieve higher results, TRA undoubtedly provides a more scientific measure of the true effectiveness of operator recognition. For instance, prediction-label pair like $[1,0,0,0,0\ldots]_{20}$ and $[0,1,0,0,0\ldots]_{20}$ will yield THA with 18/20 = 0.9, but from the operator functional perspective, it is entirely incorrect.

As for operator attribute recovery and model reconstruction, \toolname{} reaches 100\% ARA and MIA on all models in TVM; \toolname{} reaches 100\% ARA besides Inceptionv1\_2020, ShuffleNetv2\_2021, EfficientNet\_{2020} and EfficientNet\_{2021} in GLOW. The very few errors in operator attribute recovery are due to occasional wrongs of \textit{Lrn}, \textit{Avgpool} and \textit{Clip}. The attributes of these wrong operators are directly evident in the decompiled code, making it easy to correct them through manual verification. As for MIA in GLOW, one specific situation needs to be addressed: the inference confidence scores of recovered Inceptionv1 by \toolname{} differ from the executables. When ignoring these confidence score differences, the class prediction accuracy is 96.4\%. The differences in confidence are primarily attributed to precision loss during compilation. Compared to inference results of source high-level ONNX models, the MIA reaches 100\%.
After correcting all error prediction operators, both ARA and MIA of BTD for all models can achieve 100\%. However, it comes at the cost of significant time overhead. We will discuss this issue in detail in RQ2.

\begin{RQBox}
\noindent\textbf{Answer to RQ1:} \toolname{} can decompile DNN executables analyzed in BTD under different optimization levels and different compiler versions. \toolname{} overcomes BTD's inherent shortcomings in operator type recognition.
\end{RQBox}

\subsection{RQ2: Efficiency}
To answer RQ2, we compare the overhead of \toolname{} with BTD and analyze the underlying reasons.


We choose four models: EfficientNet, ResNet18, Inceptionv1 and ShuffleNetv2. These models cover a range of weights size and topological complexities, enabling a comprehensive evaluation of the overhead associated with BTD and \toolname{}. The model reconstruction strategies for \toolname{} and BTD are identical. Therefore, we only compare the time associated with operator type recognition and operator attribute recovery processes. \toolname{} involves interaction with LLM, which leads to non-fixed time consumption. We measure the time overhead by recording it three times and calculating the average to ensure a fair comparison. BTD uses IDA~\cite{ida2025} to decompile the executables and \toolname{} uses Ghidra. We overlook the difference from general decompiler preprocessing for fair comparison. The compiler version for our experiment is TVM v0.9dev and GLOW 2022. The overhead of BTD and \toolname{} is shown in Table~\ref{tab:overhead}. 
 
\begin{table*}[t]
\centering
\caption{Overhead Analysis on BTD and \toolname{} (all value in s) }
\label{tab:overhead}
\setlength{\tabcolsep}{3.5pt}
\begin{tabular}{c *{7}{cc}  }
\toprule
\multirow{2}{*}{\raisebox{-0.5\height}{\textbf{Method}}} & 
\multicolumn{3}{c}{\textbf{EfficientNet}} & 
\multicolumn{3}{c}{\textbf{ResNet18}} &
\multicolumn{3}{c}{\textbf{Inceptionv1}} &
\multicolumn{3}{c}{\textbf{ShuffleNetv2}} &
\\
\cmidrule(lr){2-4} \cmidrule(lr){5-7}\cmidrule(lr){8-10}  \cmidrule(lr){11-13}
& TVM\_O0 & TVM\_O3 & GLOW & TVM\_O0 & TVM\_O3 & GLOW & TVM\_O0 & TVM\_O3 & GLOW & TVM\_O0 & TVM\_O3 & GLOW \\
\midrule
BTD

& 676.3 & 265.0 & 2904.7 & 468.7 & 681.2 &2416.3&982.8 &705.6&4328.8& 152.7&92.8&296.2\\ 
\toolname{}

& 76.9 & 122.2 & 204.5& 41.4& 127.9& 129.2 &208.3 &288.8 &304.7&66.0& 80.9& 75.4\\
\bottomrule
\end{tabular}
\end{table*}
For TVM\_O0, TVM\_O3 and GLOW, the average time spent by BTD is about \textbf{6.79} times, \textbf{2.77} times, and \textbf{12.76} times that of \toolname{} respectively. The main time overhead for \toolname{} comes from network request to LLM and dynamic memory monitor. The time of LLM requests can fluctuate due to network conditions. However, in our implementation, requests to LLM are executed through a single thread. Using a multi-threaded approach could easily optimize the time overhead. EfficientNet represents high-capacity and computationally intensive models, while ShuffleNetv2 serves as an example of lightweight model. ResNet18 and InceptionV1 are further included to encompass a broader range of distinct architectural designs. According to our evaluation results, \toolname{} performs better than BTD in time overhead obviously across all these various models.

Moreover, it is noteworthy that \toolname{}'s approach does not involve heavy analysis constrained by hardware resources. In contrast, the methods utilized by BTD demand considerable memory and CPU resources, which can lead to performance degradation on consumer-grade devices.

\begin{RQBox}
\noindent\textbf{Answer to RQ2:} \toolname{} can decompile DNN executables with a shorter time overhead than SOTA methods and \toolname{} does not rely heavily on hardware resources.
\end{RQBox}

\subsection{RQ3: Comprehensiveness}

To answer RQ3, we aim to evaluate \toolname{} on a wider range of models and on aarch64 architecture to demonstrate its versatility. We also discuss the compatibility with quantized compiled models of \toolname{}.

We first evaluate \toolname{} on the latest compiler verison to verify its scalability. According to our observations, TVM is a project that is frequently maintained; GLOW is a stable project, we have counted the commits since 2023, which total only about 100, and the majority are related to feature maintenance and bug fixes. The repository of GLOW has been marked as public archive now. Therefore, we also use \toolname{} to analyze the latest TVM version (v0.17). We utilize \toolname{} to decompile RQ1's six models in TVM 0.17 repeatedly. We evaluate \toolname{} on six more different models in TVM v0.17, aiming to cover more model structures and varying hyperparameters for same model structure. The results are shown in Table~\ref{tab:RQ2-latest}.
\begin{table}[t]
\centering
\caption{Evaluation of \toolname on the Recent Compiler Version (all value in \%)}
\label{tab:RQ2-latest}
\setlength{\tabcolsep}{7.2pt}
\begin{tabular}{c *{3}{cc} c }
\toprule
\multirow{2}{*}{\raisebox{-0.5\height}{\textbf{Model}}} & 
\multicolumn{3}{c}{\textbf{O0}} & 
\multicolumn{3}{c}{\textbf{O3}}  \\
\cmidrule(lr){2-4} \cmidrule(lr){5-7} 
& TRA & ARA & MIA & TRA & ARA & MIA \\
\midrule
\textbf{EfficientNet}

& 100 & 100 & 100 & 100 & 100 & 100 \\ 
\textbf{Inceptionv1} 

& 100 & 100 & 100 & 97.47 & 100 & 100\\ 
\textbf{MobileNetv2} 

& 100 & 100 & 100 & 100 & 100 & 100  \\ 
\textbf{ResNet18}

& 100 & 100 & 100 & 100 & 100 & 100  \\ 
\textbf{ShuffleNetv2}

& 98.9 & 100 & 100 & 100 & 100 & 100  \\ 
\textbf{VGG16}

& 100 & 100 & 100 & 100 & 100 & 100  \\ 

\textbf{MnasNet}

& 99.49 & 100 & 100 & 100 & 100 & 100 \\ 
\textbf{ResNet34} 

& 100 & 100 & 100 & 100 & 100 & 100\\ 
\textbf{ShuffleNetv1} 

& 100 & 100 & 100 & 97.5 & 100 & 100  \\ 
\textbf{SqueezeNet}

& 100 & 97.73 & 100 & 100 & 97.85 & 100 \\ 
\textbf{Emotion}

& 100 & 100 & 100 & 100 & 100 & 100  \\ 
\textbf{SuperRes}

& 100 & 100 & 100 & 100 & 100 & 100  \\ 
\bottomrule
\end{tabular}
\end{table}

\begin{table}[t]
\centering
\caption{Evaluation of \toolname on AArch64 Models (all value in \%)}
\label{tab:RQ3-aarch64}
\setlength{\tabcolsep}{7.2pt}
\begin{tabular}{c *{3}{cc} c }
\toprule
\multirow{2}{*}{\raisebox{-0.5\height}{\textbf{Model}}} & 
\multicolumn{3}{c}{\textbf{O0}} & 
\multicolumn{3}{c}{\textbf{O3}}  \\
\cmidrule(lr){2-4} \cmidrule(lr){5-7} 
& TRA & ARA & MIA & TRA & ARA & MIA \\
\midrule
\textbf{ShuffleNetv1}

& 100 & 100 & 100 & 100 & 100 & 100 \\ 
\textbf{ShuffleNetv2} 

& 98.9 & 100 & 100 & 100 & 100 & 100\\ 
\textbf{MobileNetv2} 

& 100 & 100 & 100 & 93.94 & 100 & 100  \\ 
\textbf{Inceptionv1}

& 100 & 100 & 100 & 98.65 & 100 & 100  \\ 
\textbf{MnasNet}

& 98.97 & 100 & 100 & 100 & 100 & 100  \\ 
\bottomrule
\end{tabular}
\end{table}

\begin{figure}[t]
    \centering
    \includegraphics[width=\linewidth]{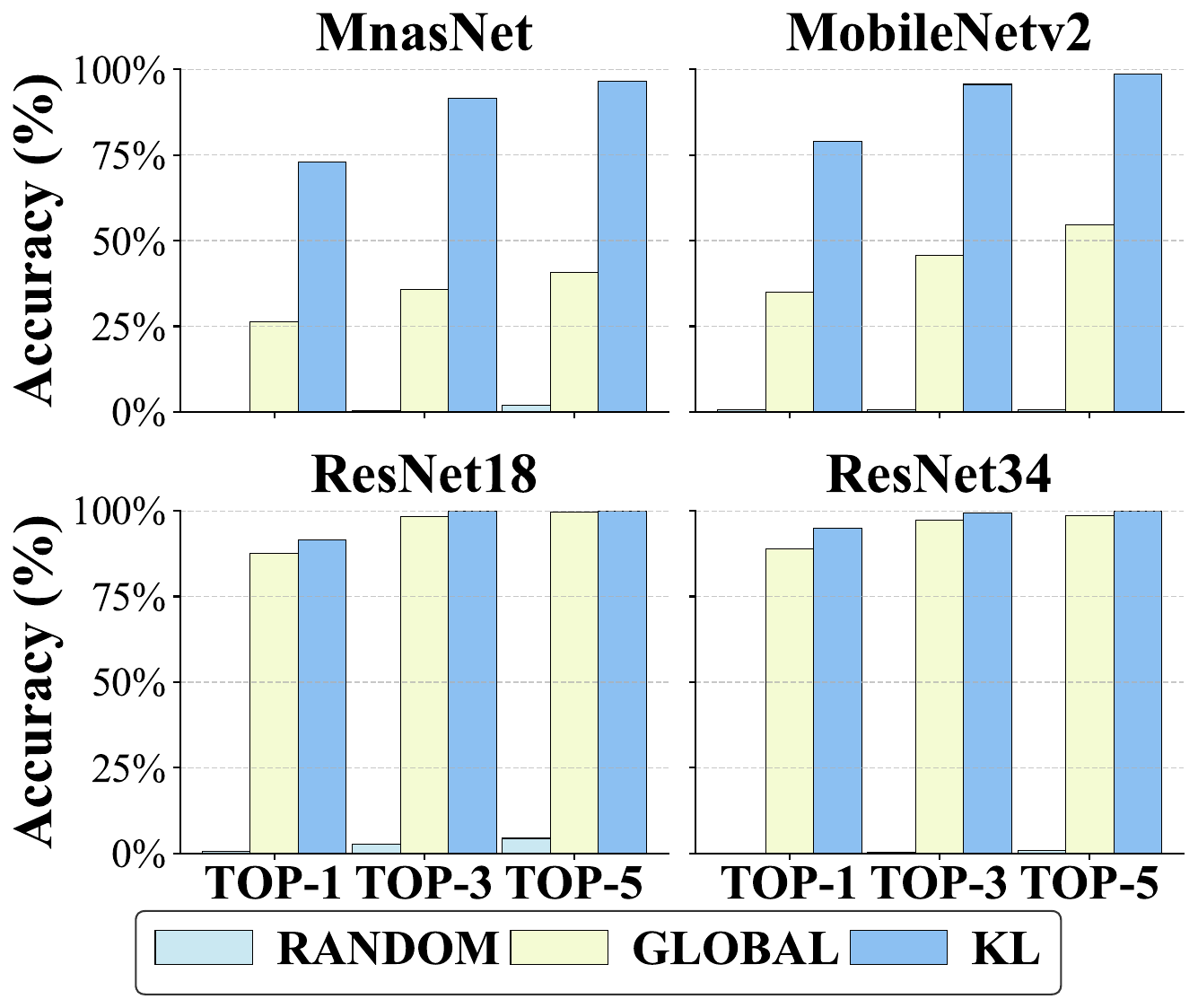}
    \caption{Evaluation of \toolname{} on Quantized Compiled Models}
    \label{fig:quant}
\end{figure}
\toolname{} can recover nearly identical high-level models for these 12 models across different optimization levels. In operator type recognition and operator attribute recovery, occasional errors may occur. The former primarily involves errors such as 
\textit{Avgpool} is incorrectly identified as \textit{Sum}. The latter involves errors in the recognition of kernel\_size for the \textit{Avgpool} , where the kernel\_size ``*0.00591716 (1/169)" should be 13 but is incorrectly identified as 17. After fixing these errors, the MIA of all models reach 100\%.

In order to show the cross-architecture support of \toolname{}, we also evaluate \toolname{} on aarch64 in TVM v0.17, the results are shown in Table~\ref{tab:RQ3-aarch64}. Based on the experimental results, \toolname{} can successfully decompile executables across different optimization levels on aarch64 architecture. Some errors may accur in operator type recognition stage. The errors include the incorrect identification of \textit{Avgpool} as \textit{Sum}, as well as misjudgments of activation functions. 

As for the evaluation of quantized compiled models, we quantized compile four models including MnasNet, MobileNetv2, ResNet18, ResNet34 using both the global\_scale and kl\_divergence methods in TVM, resulting in eight executables. We decompile these executables using \toolname{}, the results are shown in Figure~\ref{fig:quant}. 

For global\_scale method, \toolname{} can decompile all the four executables with MIA\_top1 at least 26.4\%, and MIA\_top5 at least 40.6\%. For kl\_divergence method, \toolname can decompile all the four executables with MIA\_top1 at least 73.0\%, and MIA\_top5 at least 96.4\%. Given that all models have 1000 output categories, the recovery results are quite satisfactory. In contrast, using the same computational graph with target model and random training with 5000 images (same query dataset with \toolname{}), the recovered model achieves a maximum MIA\_top1 of only 0.6\% and a maximum MIA\_top5 of only 4.4\%, indicating that it almost fails to recover any valuable information. For ResNet18 and ResNet34, \toolname{} can recover quite similar high-level models under both global\_scale and kl\_divergence methods. For MnasNet and MobileNetv2, the accuracy of recovered models decreases, we analyze the reasons as follows: compared to ResNet18 and ResNet34, MnasNet and MobileNetv2 have fewer parameters, which makes them more sensitive and difficult to determine the value of global\_scale. The global\_scale method tends to weaken the decision boundaries of models and cause the fluctuations in the model's confidence. The queries for recovered model are more likely to be classified into different categories than original high-level models. This is due to the characteristics of the global\_scale method and the inevitable precision loss during conversion between int and float domains, rather than design flaws in \toolname{}.

\begin{RQBox}
\noindent\textbf{Answer to RQ3:} We evaluate \toolname{} on a wider range of models and aarch64 architecture. Results illustrates \toolname{}'s adaptability to various models and different architectures; \toolname{} can decompile quantized compiled DNN executables and the recovered models are highly similar in functionality to the original models.
\end{RQBox}

\begin{table*}[t]
\centering
\caption{TRA of Different LLMs (all value in \%)}
\label{tab:llms}
\setlength{\tabcolsep}{6.0pt}
\begin{tabular}{c *{7}{cc}  }
\toprule
\multirow{2}{*}{\raisebox{-0.5\height}{\textbf{LLM}}} & 
\multicolumn{3}{c}{\textbf{Inceptionv1}} & 
\multicolumn{3}{c}{\textbf{ResNet18}} &
\multicolumn{3}{c}{\textbf{MobileNetv2}} &
\\
\cmidrule(lr){2-4} \cmidrule(lr){5-7}\cmidrule(lr){8-10} 
& TVM\_O0 & TVM\_O3 & GLOW & TVM\_O0 & TVM\_O3 & GLOW & TVM\_O0 & TVM\_O3 & GLOW \\
\midrule
Deepseekv3

& 100 & 100 & 96.12 & 100 & 93.94 &97.14 &100 &97.06 & 100\\ 
GPT-4.1

& 100 & 100 & 98.06& 100& 100& 100 &100&97.06& 97.67\\
Gemini2.5 flash

& 99.29 & 97.47 &86.41& 100& 100& 100 &98.57 &100& 81.40\\
GPT-4o mini

& 97.87 & 81.01 &36.89& 98.31& 87.88& 31.43 &100 &94.12& 34.88\\
\bottomrule
\end{tabular}
\end{table*}
\subsection{RQ4: Robustness}
To answer RQ4, we classify the error cases encountered by \toolname{} and introduce specific strategies to fix each type. We also evaluate \toolname{} on more LLMs.

In operator type recognition, \toolname{} may occasionally encounter errors. Additionally, due to the inherent output variability of LLMs, repeated analysis might yield different error samples. Nonetheless, our evaluation indicates that the operator type recognition accuracy on all TVM operators reach 99.22\%, and it is 97.62\% for GLOW, maintaining high accuracy. In operator attribute recovery, the errors are very rare and they are obvious in the decompiled code, such as kernel\_size “*0.00591716
(1/169)'' should be 13 but is incorrectly identified as 17. Therefore, it does not require a systematic error fix mechanism.
We analyze the causes of errors in operator type recognition and the errors can be classified into three categories. \textbf{Type1:} a single operator is split into multiple functions; \textbf{Type2:} incorrect recognition of activation functions; \textbf{Type3:} other random operator recognition errors.

The proportions of these three wrong types are 36.2\%, 13.8\% and 50\% respectively. \toolname{} has corresponding error fix strategies for different types of errors:

\textbf{Type 1: }Error operators exhibit fixed patterns. These errors accur in \textit{Softmax}, \textit{Avgpool} and \textit{Dense\_add}. These errors can be detected immediately after operator type recognition. \toolname{} can fix them by matching predefined patterns of operator split.

\textbf{Type 2: }These errors can be handled during model reconstruction. By comparing the value of intermediate layer in high-level models with those from instrumented executables, \toolname{} can locate the different operators and update the activation function according to value difference. After fixing the activation function, \toolname{} can continuously record the output value of the error operator until it matches the expected value of the executable.

\textbf{Type 3: }These errors can directly lead to crash of model construction. We can observe the exception information of the high-level model to locate error operators. \toolname{} can locate error operators and fix them by manually checking the decompiled code. After fixing the error operator, \toolname{} can continuously record the output value of error operator until it matches the expected value of executable.

To demonstrate the effectiveness of \toolname{} on different LLMs, we also compare the performance of different LLMs including GPT-4.1~\cite{GPT-41}, Deepseekv3~\cite{liu2024deepseek}, GPT-4o mini~\cite{GPT-4omini}, Gemini2.5 flash~\cite{gemini}. The selection of LLMs will affect TRA directly. Results are shown in Table~\ref{tab:llms}. Deepseekv3 and GPT-4.1 perform well in all models across different compiler settings. Gemini2.5 flash struggles to identify operators of GLOW, GPT-4o mini fails to identify TVM optimization operators and GLOW operators. In summary, \toolname does not rely on a specific LLM. LLMs that perform well in general domains are equally suitable for completing the operator type recognition task.

\begin{RQBox}
\noindent\textbf{Answer to RQ4:} \toolname{} has a stable error-fix mechanism that effectively addresses different types of errors. \toolname{} is not sensitive to the selection of LLMs, and various different LLMs are suitable for \toolname{}.
\end{RQBox}

\section{Discussion}
\subsection{Possible Defenses towards \toolname{}}
\toolname{} targets standard executables compiled using DL compilers and is capable of recovering nearly identical high-level models. Techniques such as binary obfuscation significantly reduce the readability of decompiled code by general-purpose decompilers like Ghidra. This can make \toolname{}'s operator type recognition and operator attribute recovery components ineffective. However, DL compilers do not offer direct support for executable obfuscation and standardized solutions for obfuscating DNN executables have not been established. Methods like control flow obfuscation may also negatively impact the speed and accuracy of model inference. Overall, \toolname{} is designed to handle the most common scenarios effectively.
The threat model used in our study is consistent with previous works~\cite{shi2024research,liu2023decompiling,wu2022dnd,wuneuroscope} and is generally common and practical in real-world scenarios. 
\subsection{Scalability of NLP Models}
Related studies~\cite{sun2021mind,deng2022understanding} indicate that CV models are the majority models on edge devices, and our evaluation focuses on CV models. In contrast, language models have larger search spaces of input and output, as well as greater parameters. To the best of our knowledge, there is currently no mature solution for compiling them into standalone executables in binary format  for inference on edge devices. Existing inference engines of LLMs primarily focus on optimizing the computation process but model weights are typically stored and loaded separately. Transformer-based models feature complex operator topologies, and operations like attention are decomposed into several low-level operators, further increasing the complexity of the models. This complexity makes it exceedingly difficult to precisely reconstruct an identical model. \toolname{}'s core components are still effective for NLP models, but adapting them for NLP models requires a significant amount of work including heuristic methods, we will address this in the future work.

\section{Related Work}
\subsection{LLM for Reverse Engineering}
Recent researches have demonstrated the significant potential of LLMs in reverse engineering, and there are benchmarks available to evaluate their effectiveness\cite{shang2024far,manuel2024enhancing,pordanesh2024exploring}. DecGPT~\cite{wong2023refining} leverages LLMs to repair errors in decompiled code, facilitating automatic recompilation. Lmpa~\cite{xu2023lmpa} and Degpt~\cite{hu2024degpt} combine LLMs with program analysis to optimize decompiler's output, restoring more meaningful variable names. ReSym~\cite{xie2024resym} and TypeForge~\cite{wang2025typeforge} utilize LLMs to recover user-defined structures, enhancing the readability of complex types in decompiled code. LLM4Decompile~\cite{tan2024llm4decompile} addresses the limitations of general decompiler and designs an  end-to-end decompiler using LLMs. WaDec~\cite{she2024wadec} employs a fine-tuned LLM to interpret and decompile WebAssembly, producing outputs that are more effective than general decompilers. Unlike previous works, \toolname{} focuses on analyzing DNN executables, utilizing the code understanding capabilities of LLMs to perform tasks such as operator type recognition.

\subsection{On-device Models Security}
Many previous works discuss the security of on-device models~\cite{nayan2024sok}. Huang et al. design a grey-box adversarial attack framework by comparing deep learning models on Android devices with those from TensorFlow Hub~\cite{huang2021robustness,huang2022smart}. Sun et al.~\cite{sun2021mind} discover that 41\% of machine learning apps do not protect their models at all, and even among those that employ model protection or encryption, 66\% can still be extracted. AdvDroid~\cite{deng2022understanding} conducts the first systematic study of adversarial attacks on real-world DNN models, revealing that on-device models are also vulnerable to adversarial attacks. Hu et al.~\cite{hu2023first} perform an empirical study on on-device models in iOS apps, uncovering security risks associated with deep learning models in the iOS environment. DeMistify~\cite{ren2024demistify} can automatically extract on-device models from mobile apps and reuse associated services, successfully executing 82.73\% of the models. Many works~\cite{wu2022dnd,liu2023decompiling,zhang2023libsteal} attempt to decompile on-device models compiled by DL compilers, this is also the threat model of \toolname{}.

\section{Conclusion}
In this work, we design \toolname{} to provide diverse support in decompiling DNN executables. \toolname{} recovers DNN executables back into high-level models through operator type recognition, operator attribute recovery and model reconstruction. \toolname{} leverages the semantic understanding capabilities of LLMs along with dynamic analysis to construct a comprehensive and robust decompilation pipeline. Our evaluations demonstrate that \toolname{} can successfully decompile DNN executables across different DL compiler settings, different architectures and quantized compiled models.

\bibliographystyle{IEEEtran}
\bibliography{main.bib}



\end{document}